\documentclass{article}

\usepackage[nonatbib, preprint]{neurips_2023}

\usepackage[utf8]{inputenc} 
\usepackage[T1]{fontenc}    
\usepackage{url}            
\usepackage{booktabs}       
\usepackage{amsfonts}       
\usepackage{nicefrac}       
\usepackage{microtype}      
\usepackage{times}
\usepackage{epsfig}
\usepackage{graphicx}
\usepackage{amsmath}
\usepackage{amssymb}
\usepackage{graphicx}
\usepackage{amsmath}
\usepackage{amssymb}
\usepackage{booktabs}
\usepackage{float}
\usepackage{bbm}
\usepackage[table]{xcolor}
\usepackage{xcolor}
\usepackage{colortbl}
\usepackage{multirow}
\usepackage{mmstyles}
\usepackage{multicol}
\usepackage{bbm}
\usepackage{algorithm}
\usepackage{algpseudocode}
\definecolor{myy}{RGB}{126,95,0}
\definecolor{mygray}{gray}{.9}
\definecolor{mygray2}{gray}{0.5}
\definecolor{bblue}{RGB}{30,80,120}
\definecolor{mygray1}{gray}{.7}
\definecolor{codeblue}{rgb}{0.25,0.5,0.5}
\definecolor{ggray}{RGB}{127,127,127}
\definecolor{mygreen}{RGB}{93,174,86}
\definecolor{darkergreen}{RGB}{21, 152, 56}
\definecolor{red2}{RGB}{252, 54, 65}
\definecolor{y}{RGB}{255,230,0}
\definecolor{p}{RGB}{236,185,255}
\definecolor{g}{RGB}{0,235,0}
\definecolor{textcolor1}{rgb}{0.25,0.5,0.5}
\definecolor{textcolor2}{rgb}{0.7,0.25,0.25}
\usepackage{hhline}
\usepackage{booktabs}
\usepackage{algorithm}
\usepackage{algpseudocode}
\usepackage{subcaption}
\usepackage{pifont}
\usepackage{amsfonts}
\usepackage{amsmath}
\usepackage{arydshln}
\usepackage{multirow}
\usepackage{multicol}
\usepackage[export]{adjustbox}
\usepackage[normalem]{ulem}
\usepackage{xspace}
\usepackage{rotating}
\usepackage{xhfill}
\usepackage{ulem}
\usepackage{CJKulem}
\usepackage{soul}
\usepackage{fbox}
\usepackage{wrapfig}
\usepackage{tcolorbox}
\usepackage{listings}
\usepackage{xcolor}
\definecolor{codegreen}{rgb}{0,0.6,0}
\definecolor{codegray}{rgb}{0.5,0.5,0.5}
\definecolor{codepurple}{rgb}{0.58,0,0.82}
\definecolor{backcolour}{rgb}{0.95,0.95,0.92}
\definecolor{citecolor}{HTML}{0071BC}
\definecolor{linkcolor}{HTML}{ED1C24}
\definecolor{highlightcolor}{HTML}{ABCDEF}
\usepackage{url}            
\usepackage[pagebackref, breaklinks, colorlinks, letterpaper=true, citecolor=citecolor, linkcolor=linkcolor, bookmarks=false]{hyperref}

\definecolor{mediumpurple}{rgb}{0.58, 0.44, 0.86}
\usepackage{subcaption}
\lstdefinestyle{mystyle}{
    backgroundcolor=\color{backcolour},   
    commentstyle=\color{codegreen},
    keywordstyle=\color{magenta},
    numberstyle=\tiny\color{codegray},
    stringstyle=\color{codepurple},
    basicstyle=\ttfamily,
    breakatwhitespace=false,         
    breaklines=true,                 
    captionpos=b,                    
    keepspaces=true,                 
    moredelim=**[is][\color{textcolor2}]{@}{@},
    showspaces=false,                
    showstringspaces=false,
    showtabs=false,                  
    tabsize=2
}

\newcommand{\our}{\textsc{Visor}{GPT}}
\newcommand{\light}{\textsc{Visor}{GPT}$^{\dagger}$}

\title{\Large \textsc{\textcolor{citecolor}{Visor}\textcolor{mediumpurple}{GPT}}: Learning Visual Prior via Generative Pre-Training}

\author{
    Jinheng Xie$^{1}$\quad Kai Ye$^{2*}$\quad Yudong Li$^{2*}$\quad Yuexiang Li$^3$\quad Kevin Qinghong Lin$^1$\quad \\  \textbf{Yefeng Zheng$^3$\quad Linlin Shen$^2$\quad Mike Zheng Shou$^{1\dagger}$}\\
    $^1$Show Lab, National University of Singapore\quad 
    $^2$Shenzhen University\quad
    $^3$Jarvis Lab, Tencent \\
    \texttt{\{sierkinhane,mike.zheng.shou\}@gmail.com} 
\\ \vspace{-0.6em} \\ 
\url{https://sierkinhane.github.io/visor-gpt}
}

\begin{document}
    
        \maketitle
        \let\thefootnote\relax\footnotetext{$*$ Equal Contribution\quad $\dagger$ Corresponding Author
        }

    \begin{abstract}
        Various stuff and things in visual data possess specific traits, which can be learned by deep neural networks and are implicitly represented as the visual prior, \emph{e.g.,} object location and shape, in the model. Such prior potentially impacts many vision tasks. For example, in conditional image synthesis, spatial conditions failing to adhere to the prior can result in visually inaccurate synthetic results. This work aims to explicitly learn the visual prior and enable the customization of sampling. Inspired by advances in language modeling, we propose to learn \textbf{Vis}ual pri\textbf{or} via \textbf{G}enerative \textbf{P}re-\textbf{T}raining, dubbed \textsc{\textcolor{citecolor}{Visor}\textcolor{mediumpurple}{GPT}}. By discretizing visual locations of objects, \emph{e.g.,} bounding boxes, human pose, and instance masks, into sequences, \our~can model visual prior through likelihood maximization. 
        Besides, prompt engineering is investigated to unify various visual locations and enable customized sampling of sequential outputs from the learned prior. 
        Experimental results demonstrate that \our~can effectively model the visual prior, which can be employed for many vision  tasks, such as customizing accurate human pose for conditional image synthesis models like ControlNet. 
        Code is available at \url{https://github.com/Sierkinhane/VisorGPT}.
    \end{abstract}
        \vspace{-10pt}
    \section{Introduction}
\begin{figure*}[t]
        \centering
        \includegraphics[width=\linewidth]{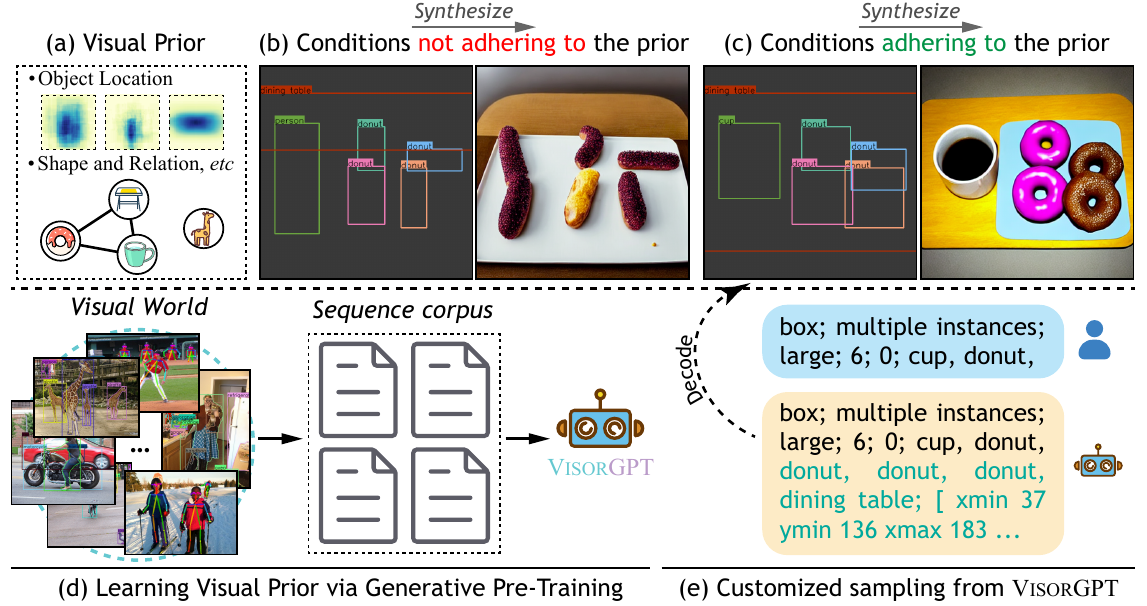}
        \vspace{-18pt}
        \caption{An overview of the problem of visual prior (top) and \our~(bottom). 
    (a) refers to visual prior, \emph{e.g.,} location, shape, and relations of objects. (b) provides a \textcolor{red}{\textit{failure}} case of image synthesis from spatial conditions that do not adhere to the prior. Specifically, the shape of the `donut' not being square and `donut' being floated in the air instead of being placed on `dining table'.
    (c) displays a \textcolor{mygreen}{\textit{success}} case that conditions sampled from \our~leads to a more accurate synthetic results. (d) illustrates that  \our~learns visual prior through sequence corpus converted from the visual world. (e) gives an example that a user customizes a sampling from \our~by prompting.
    } 
        \label{fig:teaser}
        \vspace{-20pt}
    \end{figure*}

        \vspace{-5pt}
        The digital camera can continuously capture photographs of the visual world, such that tremendous photos and videos are currently shared on the Internet. In our world, various stuff and things possess specific \textit{traits}, which have been correspondingly embedded in such visual data. In the current era of deep learning, deep neural networks~\cite{resnet,transformer,vit} have demonstrated remarkable proficiency in learning from vast amounts of data, leading to the development of visual foundation models (VFMs)~\cite{sam,sd,controlnet,composer,gligen,blip2}. Such \textit{traits} have been accordingly learned and implicitly represented as the \textit{visual prior} in VFMs, which has the potential to impact real-world applications. An example that highlights its importance can be seen in the field of image synthesis. To present high-quality and natural-looking images, the synthetic stuff and things must adhere to the visual prior such as the \textbf{spatial location, shape, and interaction of objects} (Fig.~\ref{fig:teaser} (a)). A vivid example of layout-to-image is provided in Fig.~\ref{fig:teaser} (b). When the spatial conditions do not adhere to the visual prior, such as the shape of `donut' not being square, the size of `person' being similar to that of `donut', and `donut' being floated in the air instead of being placed on `dining table', the resulting synthetic contents may be inaccurate and visually inconsistent with the desired outcome. Despite recent advances in conditional image synthesis such as ControlNet~\cite{controlnet} and GLIGEN~\cite{gligen}, the challenge of continuously sampling customized spatial conditions that adhere to the visual prior remains a difficult problem, particularly for automatic synthesis of massive images with corresponding fine-grained annotations.
  
        In this paper, we study the problem of how to explicitly learn visual prior from the real world and enable customization of sampling. If we would like to paint a series of instances on a canvas, we should decide what to paint and also their shapes, locations, interactions, \emph{etc}. It seems that these elements share a joint probabilistic prior, in which any stuff or things can be accordingly sampled to construct a scene. As there may be many potential variables in the prior, it is extremely hard to be comprehensively formulated. Over the past few years, significant advances have been made in language modeling~\cite{gpt1,gpt2,gpt3,bert}, demonstrating their remarkable capacity for modeling the probabilistic distribution of sentences. Our focus is on learning the visual prior of location, shape, and relationships among categories, rather than raw pixels. It is possible to convert such visual information into a series of sequences, such that the visual prior can be learned by language modeling. To this end, as presented in Fig.~\ref{fig:teaser} (d), we propose to learn \textbf{Vis}ual pri\textbf{or} via \textbf{G}enerative \textbf{P}re-\textbf{T}raining, dubbed \textsc{\textcolor{citecolor}{Visor}\textcolor{mediumpurple}{GPT}}. Thanks to the development of deep learning, many high-quality annotated data such as bounding-box~\cite{mscoco,objects365,openimages}, human pose~\cite{mscoco,crowdpose}, instance mask~\cite{mscoco} are publicly available. This provides sufficient location, shape, and relation information of stuff and things in the visual world. Since they are all encoded using 2D or 3D coordinates, we can simply convert them into a corpus of sequences. In this way, the visual prior can be learned by a pretext objective, \emph{e.g.,} maximizing the likelihood of each sequence. Beyond this, prompt engineering is investigated to unify various visual locations and enable the customized sampling of sequential outputs from the learned prior. 
        
        As shown in Fig.~\ref{fig:teaser} (e), according to the user's prompt, \our~can correspondingly sample a sequence from the learned prior, which can be spatially decoded for image synthesis (Fig.~\ref{fig:teaser} (c)). Since the decoded conditions adhere to the prior, the synthetic `cup', `dining table', and `donut' are realistic and consistent with the desired semantics. This finding confirms that we can continuously customize spatial conditions from many aspects, \emph{e.g.,} \textbf{data type, object size, number of instances, and classes}, using \our. With the advance of conditional image synthesis, it is feasible to generate an endless supply of synthetic images with their corresponding fine-grained annotations, potentially providing ample resources to train more robust and generalized visual intelligence models.



    
    \vspace{-5pt}
    \section{Related Works}
    \vspace{-5pt}
    \textbf{Language Modeling}. Language modeling aims to estimate the probability of a given sequence of words occurring in a sentence. 
    In recent years, transformer-based GPT series~\cite{gpt1,gpt2,gpt3} and BERT family~\cite{bert,albert,roberta} have revolutionized the field of natural language processing. In particular, BERT family adopts the encoder(-decoder) architecture and employs masked language modeling techniques to model each given sentence bi-directionally in context. In contrast, GPT series employ the decoder-only architecture to sequentially model the probability of the following tokens by maximizing the likelihood of each given sentence. Such a straightforward pretext objective allows for easy scaling up in terms of the model's parameters and training corpus.
    In this work, inspired by GPT series, we investigate the potential of a decoder-only architecture in modeling the visual probabilistic prior.
 
    \textbf{Conditional Image Synthesis}. With large-scale image-text datasets~\cite{laion400m, laion5b}, generative models, \emph{e.g.,} DALL-E~\cite{dalle,dall2}, Imagen~\cite{imagen}, and Stable Diffusion~\cite{sd}, have shown a significant capacity for synthesizing images of higher quality and greater diversity.  Recently, more controllable image synthesis models, \emph{e.g.,} ControlNet~\cite{controlnet} and GLIGEN~\cite{gligen}, have demonstrated a remarkable ability to precisely control the synthetic contents. When it comes to generating an extensive set of novel images, relying solely on spatial conditions from users or referring from images is inefficient. To tackle this problem, our~\our~is capable of continuously sampling customized and novel spatial conditions, making it possible to synthesize endless streams of data for various practical applications.
    
    \vspace{-5pt}
    \section{Methodology}
    \vspace{-5pt}
    In this section, we begin by presenting our problem formulation~($\S$~\ref{sec:pf}) and prompt designs~($\S$~\ref{sec:pe}) for unifying various visual information (\emph{e.g.,} class and location) as textual sequences. Building upon this, we introduce our model architecture and pretext objective~($\S$~\ref{sec:model}) to model the visual prior. 
    Finally, we provide practical examples of how to sample customized sequential outputs from \our~($\S$~\ref{sec:app}).
    \subsection{Problem Formulation}
    \label{sec:pf}
    We assume that visual location $\boldsymbol{x}$, \emph{e.g.,} object bounding-box, human pose, and instance mask, follow a probabilistic prior distribution $p_{\boldsymbol{x}}$. However, since $p_{\boldsymbol{x}}$ is often unavailable in practice, this work aims to learn a model $f_{\Theta}$ with parameters $\Theta$ that can empirically approximate the latent probabilistic prior $p_{\boldsymbol{x}}$. By doing so, we can sample new instances $\tilde{\boldsymbol{x}}$ from the learned prior, denoted as $\tilde{\boldsymbol{x}}\sim f_{\Theta}$, to facilitate various vision tasks, such as conditional image synthesis and action generation.
    \subsection{Visual Location as Sequences}
    \label{sec:pe}
    In the \textit{discrete} language domain, we witness that a variety of tasks, \textit{e.g.,} translation and question-answering, can be integrated as a unified template (\emph{e.g.,} prompts and responses) and then processed by one language model using different user prompts. However, when it comes to annotations of visual locations such as 2D object bounding-box, instance mask, and 3D human pose which are \textit{continuous} and highly structured, a unified approach has yet to be explored and our objective is to investigate potential solutions to this issue.
    Following Chen~\etal. \cite{pix2seq}, we discretize visual annotations, \textit{i.e.,} continuous numbers, into $m$ bins, such that location information can also be naturally represented as discrete tokens and $m$ integers are then accordingly added to the standard vocabulary. It means that coordinates can be represented by a sequence of words. 
    In particular, each number representing visual localization will be quantified as an integer in the range of $[1, m]$. In this way, visual locations $\boldsymbol{x}$ of each image can be then unified into a sequence $\mathbf{t}=\textsc{Prompt}\left(\boldsymbol{x}\right)$. 
    
    As visual annotations of various tasks are in different formats, we propose two universal prompts:
    \vspace{-5pt}
    \footnotesize
     \begin{tcolorbox}[boxrule=0pt, colframe=white, sharp corners, left=1mm, right=1mm, top=0.2mm, bottom=0.2mm]
    \fparbox[boxrule=0pt, bT]{
        \textcolor{black}{Prompt template T$_a$}:   
        
        \textcolor{textcolor2}{Annotation type; Data type; Size; \#Instances; \#Keypoints; Category names; Coordinates}
        
        \textcolor{black}{Example}: 
        
        \textcolor{textcolor1}{box; multiple instances; large; 3; 0; person, motorcycle, bicycle; [ xmin 377 ymin 250 xmax 406 ymax 288] [ xmin 287 ymin 228 xmax 377 ymax 399] [ xmin 388 ymin 258 xmax 413 ymax 286] }

        \vspace{0.5em}
        \textcolor{black}{Prompt template T$_b$}:   
        
        \textcolor{textcolor2}{Annotation type; Data type; Size; \#Instances; \#Keypoints; [Category name i Coordinate i]$_\text{i}$}
    
            \textcolor{black}{Example}: 
            
            \textcolor{textcolor1}{box; multiple instances; large;
             3; 0; [ keyboard xmin 0 ymin 268 xmax 512 ymax 384 ] [ dining table xmin 0 ymin 95 xmax 512 ymax 442 ] [ cup xmin 97 ymin 82 xmax 503 ymax 443 ]}
    }
    \end{tcolorbox}
    \vspace{-5pt}
    \begin{wraptable}{r}{6cm}
    \captionsetup{font=small}
    \scriptsize
        \centering
            \vspace{-10pt}
        \caption{Candidate choices of prompt template.}
            \vspace{-5pt}
        \begin{tabular}{llc}
            \toprule[1pt]
            Annotation type & box; keypoint; mask \\
            Data type & object centric; multiple instances\\
            Size & small; medium; large \\
            \#Instances & 1; 2; 3; $\cdots$\\
            \#Keypoints & 14; 18\\
            Category name & cup; person; dog; $\cdots$\\
            \bottomrule[1pt]
        \end{tabular}
            \vspace{-10pt}
        \label{tab:cc}
    \end{wraptable} 
    \normalsize
    The provided prompts can be summarized in Tab.~\ref{tab:cc}, which provides standardized templates to unify commonly used 2D and 3D visual location information into 1D textual sequences. Each prompt begins with the flags {\textcolor{textcolor2}{[Annotation type]}} and \textcolor{textcolor2}{[Data type]}, which are the flags indicating the type of annotation and scene, \emph{e.g.,} box and multiple instances. The following flags of [\textcolor{textcolor2}{Size]} and \textcolor{textcolor2}{[\#Instances]} represent the average area and the number of instances in the current image, while [\textcolor{textcolor2}{\#Keypoints]} indicates the number of keypoints annotated for each person, \emph{i.e.,} 14 or 18. The last two flags are the \textcolor{textcolor2}{[Category name]} of each instance and their corresponding \textcolor{textcolor2}{[Coordinate]}. We provide corresponding examples of a sequence derived from bounding-box annotations of an image. In particular, \textcolor{textcolor1}{(xmin, ymin)} and \textcolor{textcolor1}{(xmax, ymax)} are special tokens indicating the top-left and bottom-right corners of the target object. For human pose and instance mask, we use ``{a, b, c, d, $\cdots$}'' and ``{m0, m1, m2, m3, $\cdots$}'' as special tokens to distinguish each human keypoint and object boundary coordinate, respectively. Additional details are included in supplementary materials.
    For images with multiple instances, the sample order will be shuffled.
    By employing our defined templates, we transform commonly used visual annotations into a large-scale sequential corpus. The corpus can be seamlessly ingested by language models, facilitating better learning of visual commonsense prior.

    \subsection{Learning Visual Prior via Generative Pre-Training}
    \label{sec:model}
    \textbf{Model Architecture}. In the past few years, many large language models have been successively proposed, such as GPT~\cite{gpt1, gpt2, gpt3} and BERT~\cite{bert,albert,roberta} family, and recently introduced LLaMA~\cite{llama}. We employ the GPT decoder-style transformer as our model to learn the visual probabilistic prior.
    
    \textbf{Pretext Objective}. After processing the visual locations $\boldsymbol{x}$ as textual sequences $\mathbf{t}$ in $\S$~\ref{sec:pe}, we tokenize each sequence by byte-pair encoding (BPE) algorithm~\cite{bpe} to obtain a sequence with $n$ tokens $\mathbf{u} = \{u_1, u_2, \cdots, u_n\}$ such that a standard language modeling objective can be directly employed to learn visual prior by maximizing the following likelihood:
    \begin{equation}
        \mathcal{L} = \sum_{i} \text{log} p(u_i | u_{i-k}, \cdots, u_{i-1}; \Theta), 
    \end{equation} 
    where $k$ is the size of context window, and $p(\cdot | \cdot)$ indicates the conditional probability which is modeled by the neural network $\Theta$.  Stochastic gradient descent is used to train the neural network. 

    \subsection{Customizing Sequential Output}
    \label{sec:app}
\begin{figure*}[!ht]
        \centering
        \includegraphics[width=\linewidth]{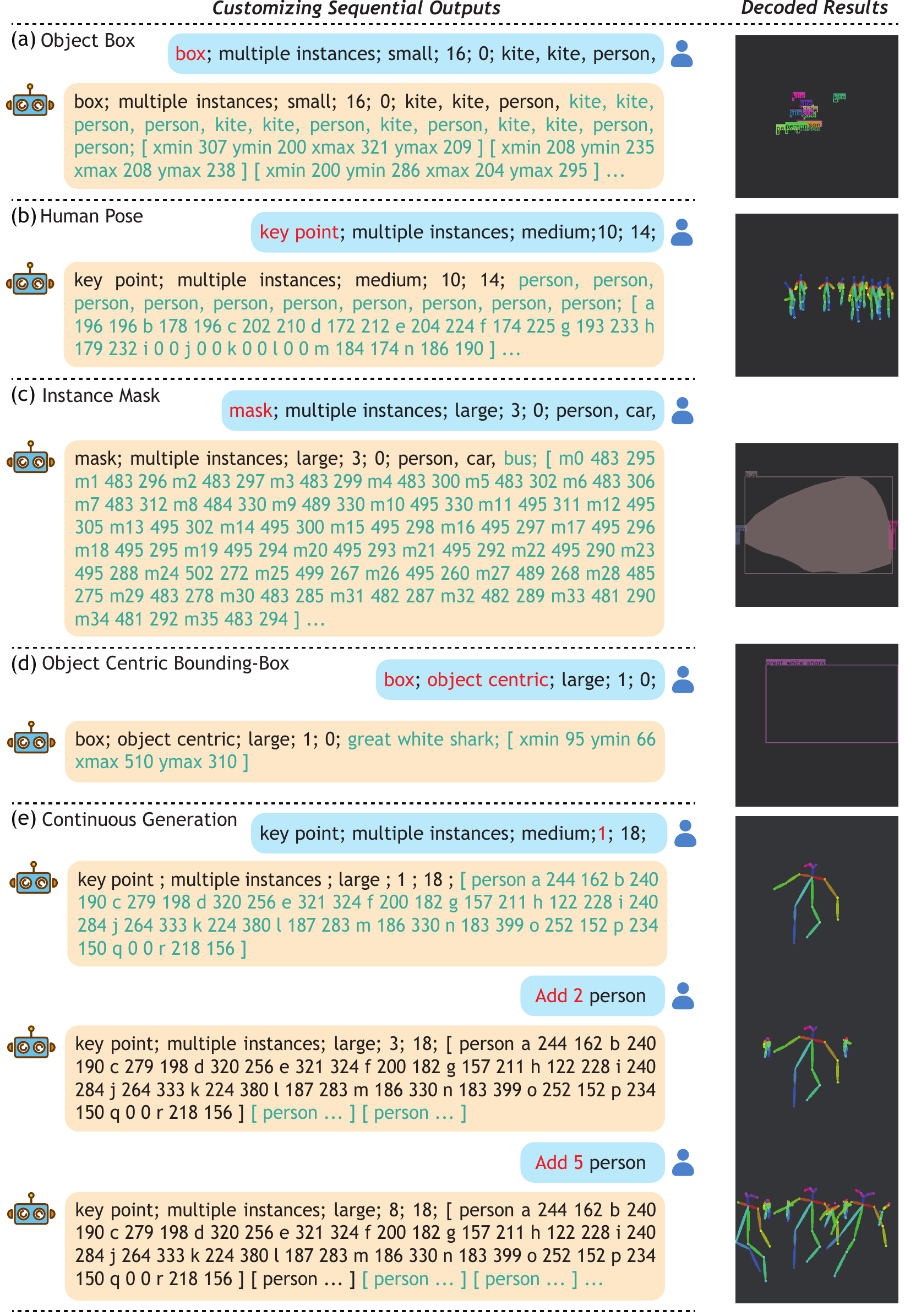}
            \vspace{-15pt}
        \caption{Examples of customizing sequential outputs from the proposed \our.} 
        \vspace{-25pt}  
            \label{fig:ctm_seq}
    \end{figure*}

    In addition to offering formatted visual annotations for learning a probabilistic prior, the standardized templates enable to \textit{personalize sequential output for various applications through prompting}. For example, the customized sequential output can be employed as spatial conditions in image synthesis models (\emph{e.g.,} ControlNet~\cite{controlnet} and GLIGEN~\cite{gligen}).  This opens up the possibility of synthesizing a broad range of data types to address diverse problems and challenges in computer vision. Here are a few representative scenarios:


    \textbf{(a) Object Bounding-Box.} As we use a flag to distinguish different types of visual annotations, we can control the type of data and scene to be sampled from the learned probabilistic prior by setting the beginning tokens in the input prompt. Accordingly, we can set the beginning prompt as ``\textcolor{purple}{box};'' to generate sequential output with instances and corresponding bounding-box information. Besides, with flags like [Size], [\#Instances], and [\#Keypoints], we can sample a scene that adheres to multiple conditions. As depicted in Fig.~\ref{fig:ctm_seq} (a), we can input a prompt ``box; multiple instances; small;  16; 0; kite, kite, person,'' as a prefix to require the \our~to conditionally infer the remaining tokens. In this example, \our~outputs the categories and their locations, specifically fulfilling the requirement of objects being in small size.

        
        
        

    \textbf{(b) Human Pose.} With flags of [\#Instances] and [\#Keypoints], \our~is capable of customizing sequential outputs involving instances with keypoints in a crowd scene. We give an example in Fig.~\ref{fig:ctm_seq} (b). Numbers (10 and 14) are added to the beginning of prompt as conditions to infer a scene consisting of 10 people with 14 keypoints. 
        
        
        
        

    
    \textbf{(c) Instance Mask.} Beyond sparse coordinates as shown in (a) and (b), \our~can deal with dense spatial annotations, \textit{i.e.,} instance masks. Typically, pixel-level information can be represented using a mask matrix or a set of boundary coordinates. For convenient sequentialization, we uniformly sample $n$ points along the angle in the polar space from object boundary coordinates to represent the pixel-level location, which is similar to \cite{polarmask}. We provide an example in Fig.~\ref{fig:ctm_seq} (c).

    \textbf{(d) Object Centric Bounding-Box.} Apart from hanlding scenes with multiple instances, the perception of object centric images~\cite{imagenet} has been extensively studied in the past decade. Thanks to the flexibility of our prompt templates, \our~can infer sequential output containing only an object and its location by setting [\#Instances] as 1. An example is shown in Fig.~\ref{fig:ctm_seq} (d).
        
        
        
        
        
        \textbf{(e) Continuous Generation.} In $\S$~\ref{sec:pe}, we introduce two prompt templates to unify the visual locations where the second one enables the continuous generation of the remaining instances. As presented in Fig.~\ref{fig:ctm_seq} (e), \our~can continuously predict the next `person' with their pose and position based on the current scene. More results are provided in the supplementary materials.
            

            

    \section{Experiments}
    
    \subsection{Experimental Setup}
    
    \begin{wraptable}{r}{9cm}
        \centering
        \vspace{-60pt}
        \caption{Details of the training corpus for \our.}
            \vspace{-5pt}
        \resizebox{\linewidth}{!}{
        \begin{tabular}{lllcc}
            \toprule[1pt]
            Datasets (type) & \#Categories & \#Images & Sampling prop. & Epochs   \\
            \midrule
            Open Images (Box)~\cite{openimages} & 600 & 1,743,042 & 33.1\% & 2.81   \\
            Objects365 (Box)~\cite{objects365} & 365 & 1,728,775 & 33.1\% & 2.81 \\
            COCO (Box)~\cite{mscoco} & 80 & 117,266 & 6.6\% & 0.56   \\
            ImageNet (Box)~\cite{imagenet} & 1,000 & 38,285 & 0.6\% & 0.06 \\
            COCO (Keypoint)~\cite{mscoco} & 1 & 53,473 & 8.3\% & 0.70 \\
            CrowdPose (Keypoint)~\cite{crowdpose} & 1 & 9,981 & 1.7\% & 0.14 \\
            COCO (Mask)~\cite{mscoco} & 80 & 117,266 & 16.6\% & 1.40  \\
            \bottomrule[1pt]
        \end{tabular}}
        \vspace{-10pt}
        \label{tab:data}
    \end{wraptable} 
 
    \textbf{Datasets}. We collect around 4 million sequences from the publicly available datasets for \our. 
    In particular, we consider three types of commonly used visual annotations, \emph{i.e.,} object bounding-box, human pose, and instance mask. In the MS-COCO dataset~\cite{mscoco}, we collect \textasciitilde118K images annotated with 80 categories and their object bounding-boxes and instance masks. For each image, all object bounding-boxes and instance masks with their category information are formatted to a sequence, respectively. Beyond that, \textasciitilde3.5 million bounding-box annotations of Objects365~\cite{objects365} and Open Images~\cite{openimages} are also converted to sequences. Other types of annotations (\emph{i.e.,} human keypoint) of MS-COCO (\textasciitilde54K) and CrowdPose (\textasciitilde10K) are also formatted to sequential data. For the object-centric scenario, we collect \textasciitilde4K sequences from ImageNet-1K~\cite{imagenet}. A summary is presented in Tab.~\ref{tab:data}.
    
    \textbf{Evaluation Metrics}. We propose to evaluate \our~from three aspects: \textbf{(i)} Evaluating the quality of sequences generated by \our. In the inference stage, as \our~predicts sequences in the format given in $\S$~\ref{sec:pe}, it is necessary to examine whether the generated sequences can be decoded into visual locations. In particular, we generate a series of sequences using \our~and calculate the accuracy whether it can be successfully decoded (termed \textbf{Format} in Table~\ref{tab:acc}) and the number of categories matches the number of locations (termed \textbf{Matching} in Table~\ref{tab:acc}). \textbf{(ii)} As discussed in $\S$~\ref{sec:pe}, we use flags, \emph{i.e.,} [Size] and [\#Instances], to indicate the average size and number of instances in the current sequence. Hence, we can control the average object size and the number of instances in the generated sequences via setting flags [Size] and [\#Instances]. Then, we can calculate the accuracy whether the object size and the number of instances in the generated sequences are consistent with the given flags to validate the performance of controllability (termed \textbf{Size} and \textbf{\#Instances}, respectively). \textbf{(iii)} Evaluating the learned probabilistic prior, \emph{i.e.,} object location, shape, and relation among categories, on the \textit{val} set of COCO, Objects365, and Open Images datasets. In this work, we propose to compare the discrete distribution of every visual prior. Specifically, to compute the \textbf{location} prior of a category, we initialize an empty canvas and convert the bounding-box of each instance of the category to a binary mask. Then, each mask is accumulated on the canvas and normalized as 2D location distribution.
    To compute the \textbf{shape} prior of a category, we calculate the ratio of width to height of each instance of the category, and estimate a discrete distribution as the shape prior. 
    To establish the \textbf{relation} prior of a category to other categories, we count the number of co-occurrences between the category and other categories and estimate a discrete distribution.
    In this way, discrete prior of each category can be computed on COCO, Objects365, and OpenImages \textit{val} sets as real one. Durring evaluation, we infer a series of sequences to compute the learned visual prior. Then we measure the similarity between learned and the real prior using the \textbf{Kullback-Leibler divergence}~\cite{kld}. All evaluation is based on object bounding-box. More details are provided in supp. 

    \begin{table}[h]
        \centering
            \vspace{-15pt}
        \caption{Model card of \our.}

        \resizebox{\linewidth}{!}{
            \begin{tabular}{l c c c c c c c c c}
                \toprule[1.5pt]
                Models & \#Parameters & \#Training data & Annotation type & Batch size & Iterations & Learning rate & Sequence length $n$ \\
                \midrule
                \our &  117M & 4M & box \& keypoint \& mask & 128 & 200K &  $5.0e^{-5}$  & 1024  \\
                \light & 117M  & 34K & box \& keypoint \& mask &  128 & 200K & $5.0e^{-5}$ & 1024 \\
                \bottomrule[1pt] 
        \end{tabular}}
        \label{tab:model_card}
            \vspace{-10pt}
    \end{table}
    \textbf{Implementation Details}. We provide training details of \our~in Tab.~\ref{tab:model_card}. 
    \our~adopted GPT-2 (base) architecture and was trained from scratch.
    We use all datasets reported in Tab.~\ref{tab:data} to train \our. As the number of training sequences on each dataset is significantly unbalanced, we re-sample each dataset according to the proportion as indicated in Tab.~\ref{tab:data} to train the \our. Open Images and Objects365 are not involved to train \light and there is no re-sampling. 
    In evaluation, each category is at least involved in \textasciitilde80 valid predicted sequences by prompting ($\S$~\ref{sec:pe}).

    \begin{table}[h]
        \centering
            \vspace{-10pt}
        \caption{Evaluation on training corpus scale and prompt templates of \our. The similarity between real probabilistic prior and the learned one is measured by KL divergence (KL Div).}
        \resizebox{\linewidth}{!}{
            \begin{tabular}{l l c c c c c c c c c c c}
                \toprule[1.5pt]
                \multirow{2.5}{*}{Models} &\multirow{2.5}{*}{Prompt}  &  \multicolumn{3}{c}{KL Div on COCO $(\downarrow)$} & \multicolumn{3}{c}{KL Div on Open Images $(\downarrow)$} &  \multicolumn{3}{c}{KL Div on Objects365 $(\downarrow)$}\\
                \cmidrule(lr){3-5}  \cmidrule(lr){6-8}  \cmidrule(lr){9-11}  
                & &  Location & Shape & Relation  & Location & Shape & Relation & Location & Shape & Relation\\
                    \midrule
                \light & T$_a$ & 1.133 &  1.483 & 0.452 & - & - & - & - & - & -   \\
                    \light & T$_a$+T$_b$ & 1.032 & 1.446 & 0.445 & - & - & - & - & - & - \\
                \our & T$_a$ & 1.212 &  1.813 & 0.561 & 0.890 & 2.775 & 3.715 & 1.969 & 1.345 & 2.790  \\
                    \our & T$_a$+T$_b$ & 1.583 & 1.710 & 0.581 & 1.007 & 2.782 & 3.888 & 1.995 & 1.377 & 2.765 \\
                \bottomrule[1pt] 
        \end{tabular}}
        \label{tab:compar_prior}
            \vspace{-15pt}
    \end{table}
    \subsection{Quantitative Results} 

    \textbf{Evaluation on Learned Visual Prior}. In Tab.~\ref{tab:compar_prior}, we present the measured similarity between real probabilistic prior and the one learned by \our~on the validation sets of COCO, Open Images, and Objects365, using KL divergence. 
    The prompt template T$_a$ and T$_a$+T$_b$ in $\S$~\ref{sec:pe}, are used for comparison. Overall, \our~T$_a$ and T$_a$+T$_b$ exhibit comparable  performance, indicating both prompt templates have comparable capability for learning visual prior. 
    

    \begin{wraptable}{r}{7cm}
        \centering
            \vspace{-12pt}
        \caption{Evaluation on customized outputs (\%).}
            \vspace{-6pt}
        \resizebox{\linewidth}{!}{
            \begin{tabular}{l c c c c c c }
                \toprule[1.5pt]
                \multirow{2.5}{*}{Datasets}  &  \multicolumn{2}{c}{Quality $(\uparrow)$} & \multicolumn{2}{c}{Controllability $(\uparrow)$}\\
                \cmidrule(lr){2-3}  \cmidrule(lr){4-5}   
                &  Format & Matching & Size & \#Instances \\
                \midrule
                COCO & 100.0 &  100.0  & 92.02 & 100.0   \\
                  Open Images &  99.97  &   99.40  &  89.35  &  98.71  \\
                    Objects365 &  99.99  &  99.94  &  91.52  &  99.78  \\

                \bottomrule[1pt] 
        \end{tabular}}
        \label{tab:acc}
            \vspace{-10pt}
    \end{wraptable} 
    
    \textbf{Evaluation on Customized Sequences}. We present the quality of generated sequences and the performance of \our's controllability in Tab.~\ref{tab:acc}. It is obvious that nearly all predicted sequences can be decoded successfully in three datasets. Additionally, in over 99\% of sequences, all instances can match their respective locations. Besides, the table shows that \our~achieves accuracies of 92.02\%, 89.35\%, and 91.52\% in controlling the average object size on COCO, Open Images, and Objects365 datasets, respectively. Furthermore, \our~can achieve an accuracy of over 98\% in controlling the number of instances across all three datasets. These findings demonstrate the strong capacity of \our~in reasoning high-quality sequences and control the object size and number of instances in the scene.

 
    \subsection{Visualization Results}

        \begin{wrapfigure}{r}{9cm}  
        \centering
            \vspace{-38pt}
        \includegraphics[width=\linewidth]{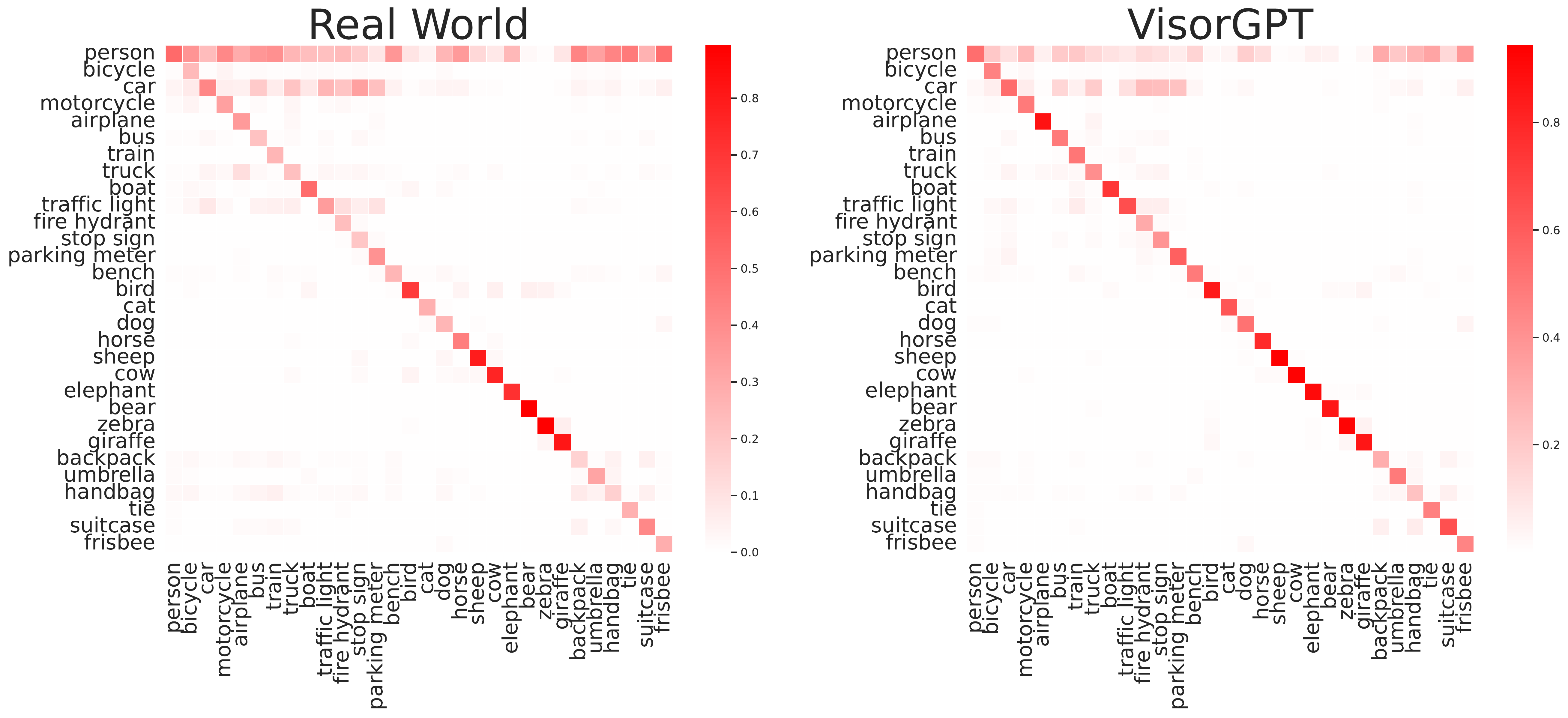}
        \vspace{-15pt}
        \caption{Relation matrix among 30 categories on COCO.}
        \label{fig:rel_priors}
        \vspace{-10pt}
    \end{wrapfigure}
    \textbf{Relation Prior.}
        Fig.~\ref{fig:rel_priors} illustrates the comparison between the real-world relation matrix among 30 categories and the one estimated by \our. Each row depicts the relation prior of one category to others. For instance, it can be observed from the real world matrix that the `person' (the first row) frequently interacts with other categories such as `dog' and `cat'. Similarly, in the third row, the co-occurrence between `car' and `bus', `truck', and `stop sign' is larger than that of other categories. 
        Notably, it is clear that the relation prior learned by \our~is very close to that of the real-world one. This indicates that \our~can capture the real relationships among categories and generate sequential output that aligns with these visual prior.
        
    \begin{wraptable}{r}{6.5cm}
            \vspace{-21pt}
        \centering
        \includegraphics[width=\linewidth]{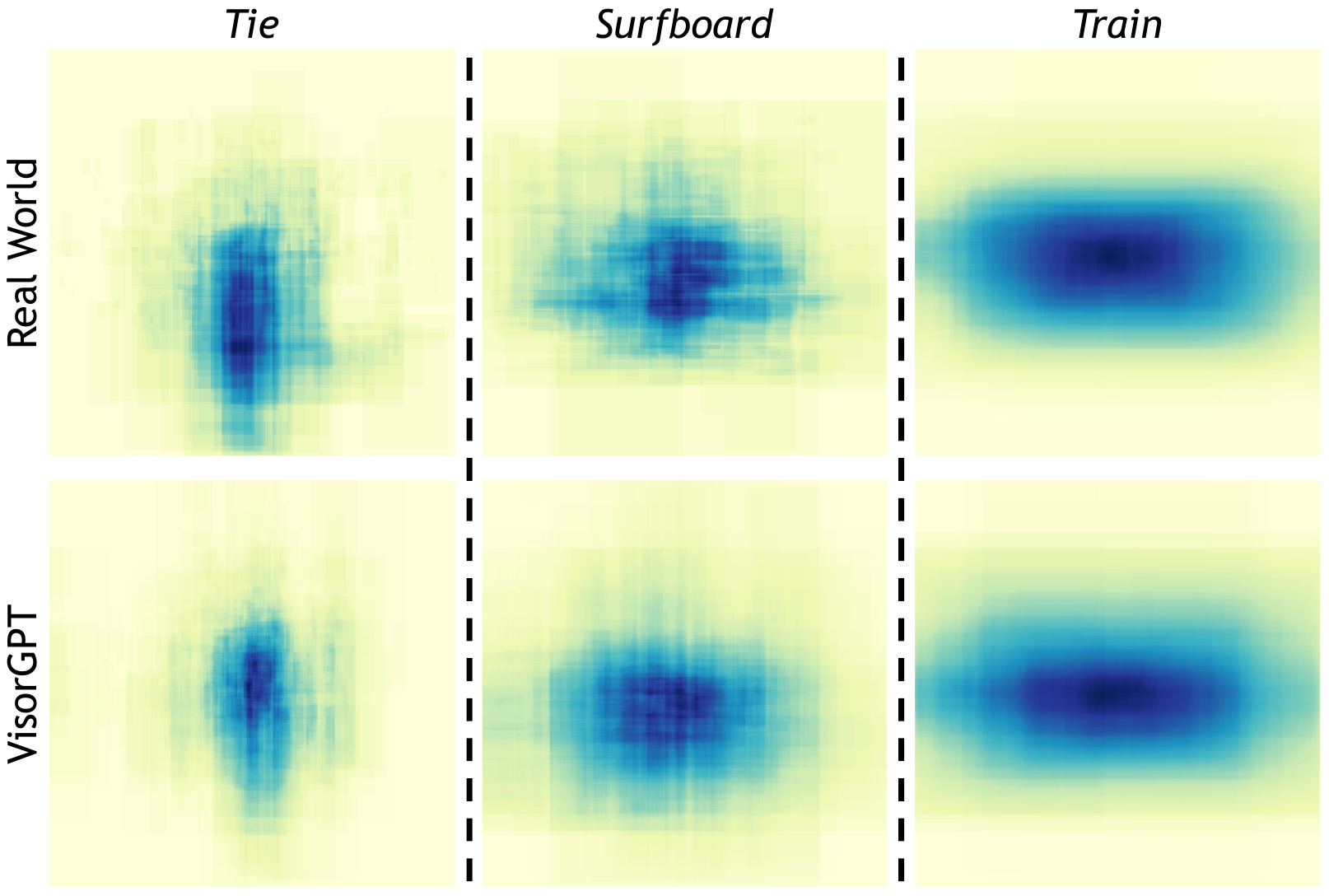}
        \vspace{-15pt}
        \caption{Location prior of some categories.} 
        \label{fig:loc_priors}
        \vspace{-15pt}
    \end{wraptable}
    
    \textbf{Location Prior.} In addition to the   
        quantitative results presented above, we visualize the comparison between the location prior learned by \our~and the real one across various categories. Fig.~\ref{fig:loc_priors} displays the location prior of three categories, including `surfboard', `tie', and `train'. It is noticeable that, in each column, the location prior learned by \our~is similar to the real one. For instance, from the first column, one can observe that the real distribution of `tie' is mainly located in the lower-middle region, and the shape prior learned by \our~exhibits a similar pattern.

\begin{figure*}[t]
        \centering
        \includegraphics[width=\linewidth]{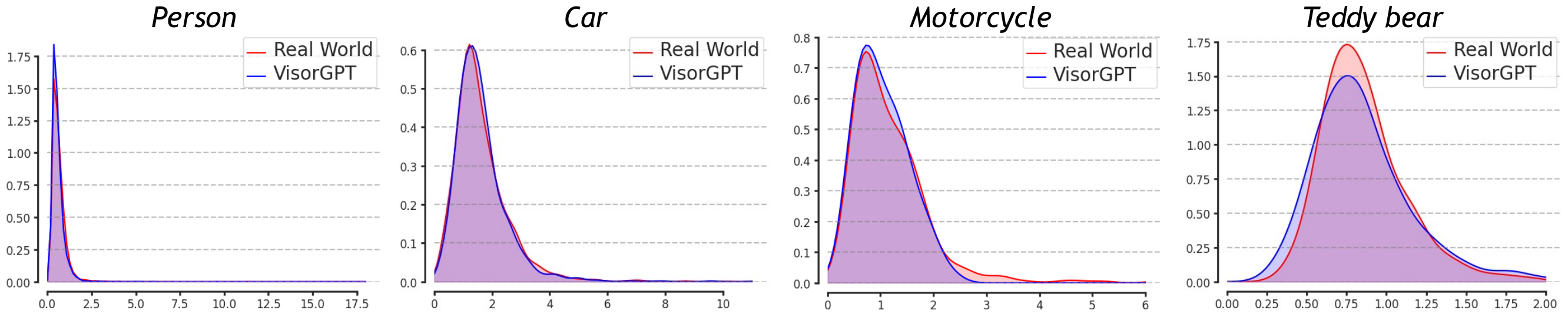}
        \vspace{-15pt}
        \caption{Shape prior of the categories of `person', `car', `motorcycle', and `teddy bear'.} 
        \label{fig:sha_priors}
        \vspace{-20pt}
    \end{figure*}

        \textbf{Shape Prior.}
        Fig.~\ref{fig:sha_priors} shows the shape prior of four categories, such as `person' and `motorcycle'. To facilitate comparison, we employ kernel density estimation to estimate a continuous distribution from the discrete one. We observe that the shape prior learned by \our~is close to those of the real visual world. For example, in the real world, the ratio of width to height of a car is almost always larger than 1, and the estimated shape prior of `car' is mainly distributed around 1.8. 
        It is evident that the learned probabilistic prior by \our, represented by the blue line, closely approximates the real one, represented by the red line. Overall, the shape priors of other categories learned by \our~well match that of the real world.
        
        \subsection{Ablation Studies}
        \vspace{-10pt}
        \begin{table}[h]
        \caption{Impact of Special Words (SW), Textual Knowledge (TK), 
        Number of Sequences (\#Seq), and Model size (\#Param). We use KL Div~$(\downarrow)$ as evaluation metric.} 
        \vspace{-4pt}
        \label{tab:ablation}
        \begin{subtable}{.4\linewidth}
        \footnotesize
          \centering
            \caption{Effect of SW and TK.}
            \vspace{-5pt}
            \label{tab:ablation_a}
            \begin{tabular}{ccccc}
                    \toprule[1pt]
                    SW & TK & COCO & Open Images \\
                    \midrule
                        \rowcolor{mygray}
                        \checkmark & $\times$ & \textbf{1.647} & \textbf{1.895}  \\
                        $\times$ & $\times$ & 1.720  & 1.950 \\
                        $\times$ & \checkmark & 1.959  & 2.240 \\
                    \bottomrule[1pt]
            \end{tabular}
        \end{subtable}%
        \begin{subtable}{.4\linewidth}
        \footnotesize
          \centering
            \caption{Effect of \#Seq.}
            \vspace{-5pt}
            \label{tab:ablation_b}
            \begin{tabular}{ccc}
                    \toprule[1pt]
                    \#Seq & COCO & Open Images \\
                    \midrule
                        \textasciitilde40 & 1.958 & 3.247  \\
                        \textasciitilde80 & 1.195  & 2.460  \\
                        \rowcolor{mygray}
                    \textasciitilde120 & \textbf{0.930} & \textbf{2.144}\\
                    \bottomrule[1pt]
                \end{tabular}
        \end{subtable} 
        \begin{subtable}{.19\linewidth}
        \footnotesize
          \centering
            \caption{Effect of \#Param.}
            \vspace{-5pt}
            \label{tab:ablation_c}
            \begin{tabular}{cc}
                    \toprule[1pt]
                    \#Param & COCO \\
                    \midrule
                        117M & 0.850  \\
                        345M & 0.836 \\
                        \rowcolor{mygray}
                        762M & \textbf{0.798} \\
                    \bottomrule[1pt]
                \end{tabular}
        \end{subtable}%
        \vspace{-10pt}
        \end{table}
    Tab.~\ref{tab:ablation} presents the impact of Special Words (SW), Textual Knowledge (TK, \emph{i.e.,} with model weights initialized from the official pre-trained GPT-2), the number of sequences (\#Seq), and model size (\#Param).
    \textbf{(a)} Results on Tab.~\ref{tab:ablation_a} are measured by the average KL divergence of location and shape prior. This confrims that the special words can potentially improve \our's performance in learning the visual prior. 
    Notably, we found that the NLP textual knowledge deteriorated the performance of \our. We attribute this to the fact that the association between visual coordinates and natural language is relatively weak, thus it becomes inessential to learn visual prior from visual annotations.
    \textbf{(b)} In Tab.~\ref{tab:ablation_b}, we find that increasing the number of sampled sequences leads to a more precise estimation of the visual prior by \our. 
    \textbf{(c)} In Tab.~\ref{tab:ablation_c}, we investigate the impact of model size on learning visual prior. For simplicity and efficiency, we replace \our~architecture by three GPT versions and train it using only COCO (box) data. The results demonstrate the scalability of \our,~\textit{i.e.,} modeling the visual prior better with increased learnable parameters.

    \begin{figure*}[b]
        \centering
        \vspace{-15pt}
        \includegraphics[width=\linewidth]{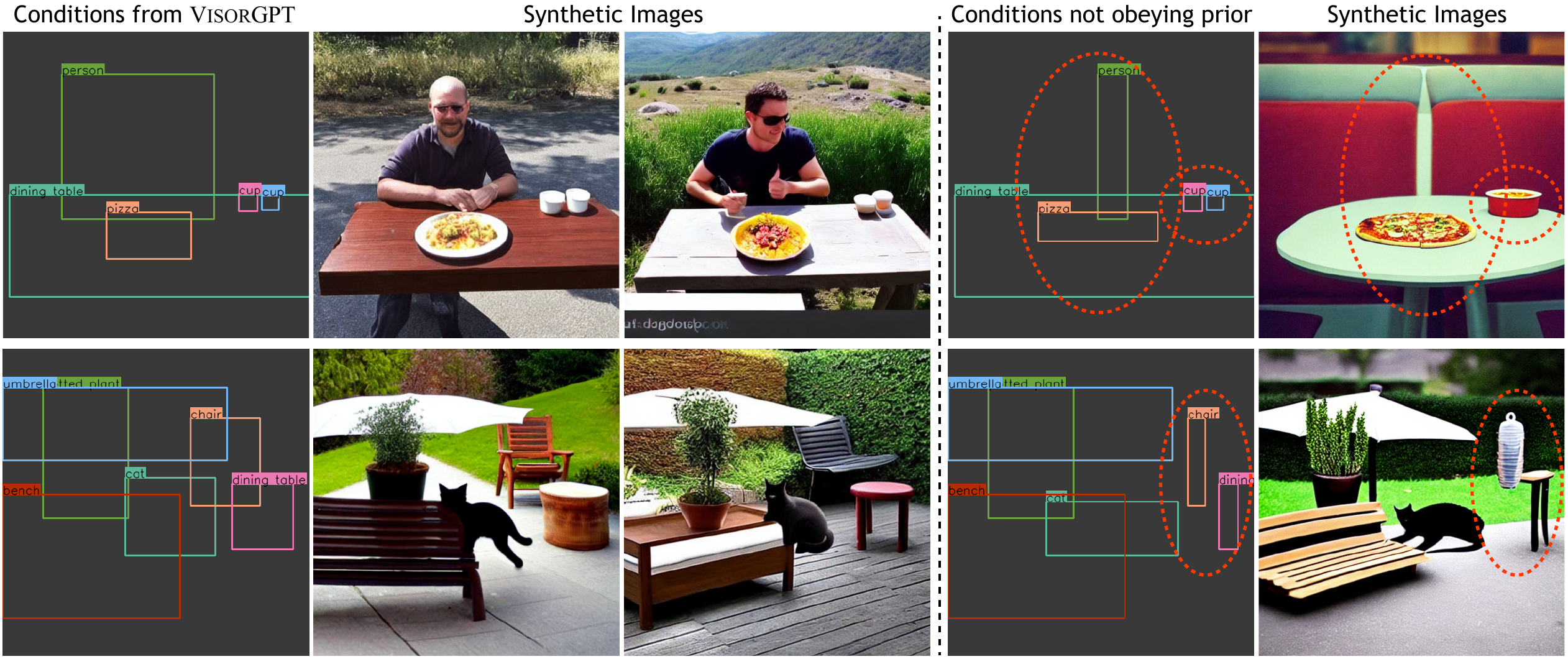}
        \vspace{-18pt}
        \caption{Comparison of synthetic images from \textit{object boxes} adhering to the prior (left) or not (right).} 
        \label{fig:seq2box}
        \vspace{-15pt}
    \end{figure*}

    \subsection{Applications}
\begin{figure*}[t]
        \centering
        \includegraphics[width=\linewidth]{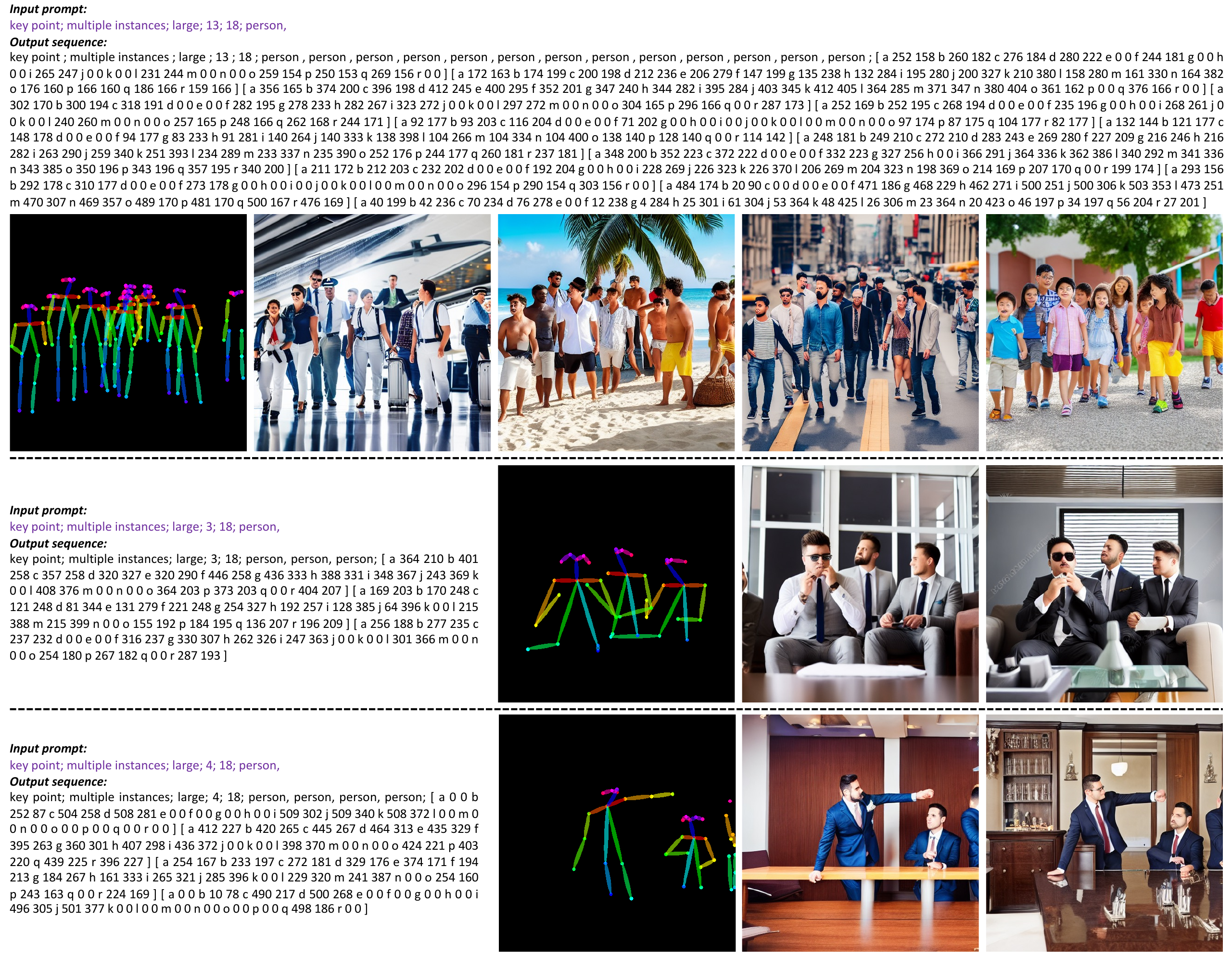}
        \vspace{-18pt}
        \caption{Illustration of input prompts (comprising multiple instances with \textit{keypoints}), output sequences, decoded results and synthetic images.} 
        \label{fig:seq2kpt}
        \vspace{-18pt}
    \end{figure*}

    \textbf{Conditional Image Synthesis}. 
    \our's~remarkable ability to infer visual categories and their locations based on user-customized prompts shows promising potential for generating customized images that still maintain a sense of realism. 
    Here, we utilize ControlNet~\cite{controlnet} and GLIGEN~\cite{gligen} to synthesize images from keypoints and bounding-boxes, respectively.
    We showcase some examples in Fig.~\ref{fig:seq2box} and \ref{fig:seq2kpt}. The first and fourth columns in Fig.~\ref{fig:seq2box} present the customized spatial conditions sampled from \our~and the conditions not adhering to the visual prior. The second, third, and fifth columns provide synthetic results by GLIGEN conditioned on the corresponding spatial conditions. For example, on the first three columns, it is evident that the spatial conditions sampled from \our~are more natural, such that the synthetic images are realistic and natural-looking. However, when the conditions (the last two columns) do not adhere to the prior, such as `person' not being on a similar scale to `dining table', the width of `pizza' being too long, and the width of `chair' being too short, the synthetic contents like `person', `chair', and `dining table' appear abnormal, also impacting the authenticity of other objects like the two cups (circled in red dotted line).
    
    Moreover, \our~is capable of inferring sequences that include instances with keypoint information. For example, as shown in Fig.~\ref{fig:seq2kpt}, we can provide a prompt like ``\textcolor{textcolor2}{key point; multiple instances; large; 13; 18; person, }'' to \our. This allows it to conditionally imagine a scene involving 13 people with their keypoint coordinates. Decoded results can be used as spatial conditions for image synthesis by ControlNet (shown in the last two columns). More examples can be found in supp.

    \section{Conclusion and Discussion}
    This work proposed a novel approach, \our, to explicitly learning the probabilistic prior of the visual world through generative pre-training. This was achieved by transforming the continuous visual locations into discrete tokens by prompting and training a transformer decoder to maximize the likelihood of training sequences. As a result, \our~exhibits significant potential in comprehending real-world visual prior and leveraging this knowledge to create plausible scenes under a variety of customized prompts.
    This ability can facilitate the automatic synthesis of a vast number of images, along with their corresponding fine-grained annotations, using ControlNet and GLIGEN. This could potentially yield ample resources to train more robust visual intelligence models.

    \section{Limitation}
    Due to the limited number of labeled classes, \our~is currently only capable of closed-set inference within approximately 1,000 categories. Additionally, we encountered limitations regarding the number of instances that could be included in each sequence due to the maximum token length, despite converting each mask annotation to a fixed length. In the future, we plan to address these limitations by incorporating natural language corpora and extending the maximum sequence length.
    
    {
        \small
        
        \bibliographystyle{ieee_fullname}
        \bibliography{ref}
    }
    

    \newpage
    \section{Appendix}
    \label{supp}
    \subsection{Examples of Training Sequences}
    Here, we give some examples of various types of training sequences on different datasets:
    \footnotesize
     \begin{tcolorbox}[boxrule=0pt, colframe=white, sharp corners, left=1mm, right=1mm, top=0.2mm, bottom=0.2mm]
    \fparbox[boxrule=0pt, bT]{
        \textcolor{black}{Human Pose (COCO)}:   
        
        \textcolor{textcolor1}{key point; multiple instances; large; 1; 18; person; [ a 190 120 b 266 146 c 318 143 d 385 232 e 338 269 f 214 150 g 0 0 h 0 0 i 312 280 j 365 296 k 359 420 l 258 283 m 194 344 n 301 383 o 197 100 p 181 103 q 234 84 r 0 0]}

        \vspace{0.5em}
        
        \textcolor{black}{Human Pose (CrowdPose)}:   
        
        \textcolor{textcolor1}{key point; multiple instances; large; 2; 14; person, person; [ a 312 201 b 306 200 c 311 232 d 269 214 e 298 257 f 231 206 g 296 275 h 307 275 i 251 244 j 271 235 k 274 292 l 283 295 m 304 153 n 310 191] [ a 179 247 b 165 245 c 164 313 d 160 315 e 221 316 f 207 279 g 155 343 h 144 366 i 242 337 j 240 367 k 210 431 l 300 418 m 172 176 n 177 227]
key point; multiple instances; large; 2; 14; person, person; [ a 240 178 b 304 168 c 228 239 d 0 0 e 261 236 f 0 0 g 251 296 h 289 296 i 0 0 j 0 0 k 0 0 l 0 0 m 261 92 n 272 156] [ a 314 160 b 363 158 c 274 232 d 356 264 e 224 260 f 271 263 g 298 315 h 341 324 i 0 0 j 332 442 k 0 0 l 0 0 m 287 64 n 333 133]}

        \vspace{0.5em}
        \textcolor{black}{Instance Mask}:   
        
            \textcolor{textcolor1}{mask; multiple instances; medium; 1; 0; clock; [ m0 224 291 m1 226 299 m2 227 306 m3 228 313 m4 233 320 m5 238 325 m6 245 329 m7 252 332 m8 259 334 m9 266 335 m10 274 333 m11 281 330 m12 288 327 m13 293 323 m14 299 318 m15 303 312 m16 305 305 m17 307 298 m18 310 291 m19 308 284 m20 307 276 m21 303 269 m22 299 263 m23 295 257 m24 288 254 m25 280 251 m26 273 250 m27 266 249 m28 259 249 m29 252 251 m30 246 256 m31 240 260 m32 235 265 m33 229 270 m34 227 277 m35 225 284]}

        \vspace{0.5em}
        \textcolor{black}{Object Centric Bounding-Box}:   
        
            \textcolor{textcolor1}{box; object centric; large; 1; 0; castle; [ xmin 236 ymin 142 xmax 413 ymax 232]}
    }
    \end{tcolorbox}
    \subsection{Implementation Details}
    All experimental evaluations were conducted on eight NVIDIA Tesla V100-32GB GPUs using PyTorch. In order to include special words, we created a new vocabulary containing a total of 30,769 words based on a standard vocabulary. To optimize computational efficiency and memory utilization, we utilized the DeepSpeed framework. To serialize visual locations, we first resized the long side of each image to a length of 512 pixels and then shifted the image content to the center by padding the short side to a length of 512 pixels. As a result, the number of bins $m$ was set to 512. The flag of \textcolor{textcolor2}{[Size]} indicates the average area of all instances in the image and we set the flag according to the rule: 
    $$ \left\{
    \begin{array}{rcl}
    \text{``small''}       &      & \text{average area} < 32^2 \\
    \text{``medium''}       &      & 32^ 2\leq \text{average area} < 96^2 \\
    \text{``large''}       &      & \text{average area} \geq 96^2 
    \end{array} \right. . $$
    We omitted person instances with fewer than five keypoints. To enable continuous generation, we designed and trained models based on the prompt format (b). Specifically, \our$^\dagger$ (a\&b) and \our~(a\&b) were trained using the same number of sequences as \our$^\dagger$ (a) and \our~(a), respectively. The only difference is that we randomly utilized prompt format (a) or (b) to construct each training sequence. 
    
    During the evaluation stage, we set the maximum sequence length of our model (\our) to 256 tokens to ensure efficient inference. In the ablation studies, we added special words only to the \textcolor{textcolor2}{[Coordinate]} term, and we reported the average KL divergence between the location and shape priors learned by \our~and those in the real world. Since training large-scale language models is time- and resource-consuming, we trained only three types of \our~with respect to GPT-2 (base, medium, large) with a maximum token length of 256 in 50,000 iterations on COCO (Box) data.
    
    \subsection{Evaluation Details}
    To estimate discrete visual prior from \our, we infer a series of sequences via prompting as below:
    
    \fparbox[boxrule=1pt, bT]{
	Code in Python:
 
	\textcolor{textcolor2}{\texttt{f"box; multiple instances; {random.choice(['small', 'medium', 'large'])}; {random.randint(2, 10)}; 0; category name,"}}
 
    }

    To ensure that each category in a given dataset is sufficiently represented in the sequence data used for estimating the visual prior, we specify a minimum number of sequences in which each category must appear. Table~\ref{tab:eval_detail} provides an overview of the predicted sequences that are used for evaluation.
    \begin{table}[h]    
		\centering
		\caption{Details about the predicted sequences for evaluation.}
		\resizebox{0.9\linewidth}{!}{
		\begin{tabular}{lccc}
			\toprule[1pt]
			Datasets & \#Categories & \#Predicted Seq. & Min \#Seq. Per Category   \\
			\midrule
			Open Images (Box) & 600 & 48,000 & \textasciitilde80    \\
			Objects365 (Box) & 365 & 29,200 & \textasciitilde80  \\
			COCO (Box) & 80 & 6,400 & \textasciitilde80   \\
			\bottomrule[1pt]
		\end{tabular}}
		\label{tab:eval_detail}
	\end{table} 

    In our study, we adopt the Kullback-Leibler divergence to quantify the similarity between two given discrete distributions. Specifically, let $p$ and $q$ denote the estimated probabilistic priors derived from the real-world data and the \our, respectively. The degree of similarity between these two distributions can be computed as:
    \begin{equation}
        \text{KL}(p||q) = p\text{log}(p/q).
    \end{equation}
    
    \subsection{Visualization}
    \begin{figure*}[h]
		\centering
		\includegraphics[width=\linewidth]{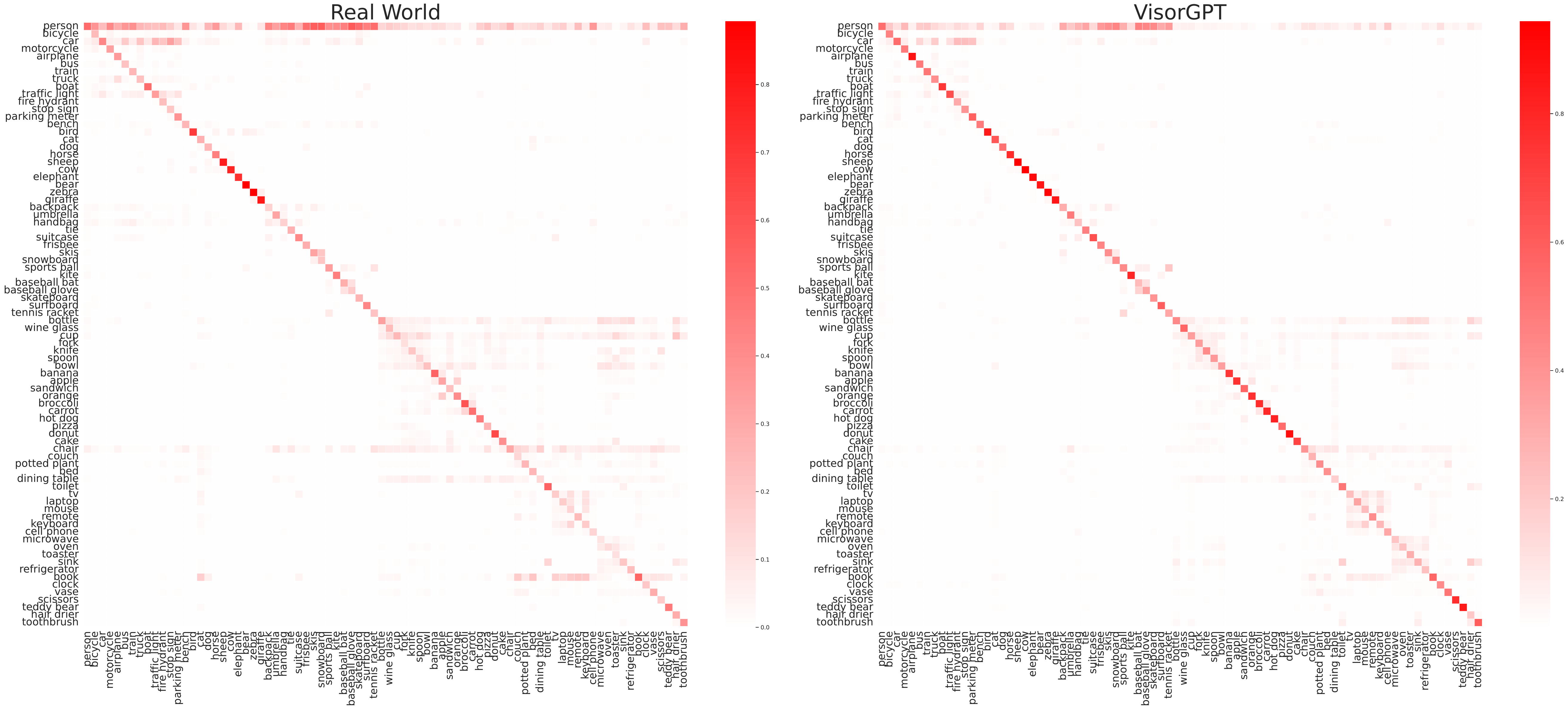}
		\caption{Relation among 80 categories on COCO.} 
		\label{fig:relation_appendix}
	\end{figure*}
    \textbf{Relation Prior of COCO}. Fig.~\ref{fig:relation_appendix} illustrates the comparison between the real and learned relation prior among 80 categories on the COCO dataset. As can be observed, there is a high degree of similarity between the two relation matrices.

\begin{figure*}[h]
		\centering
		\includegraphics[width=\linewidth]{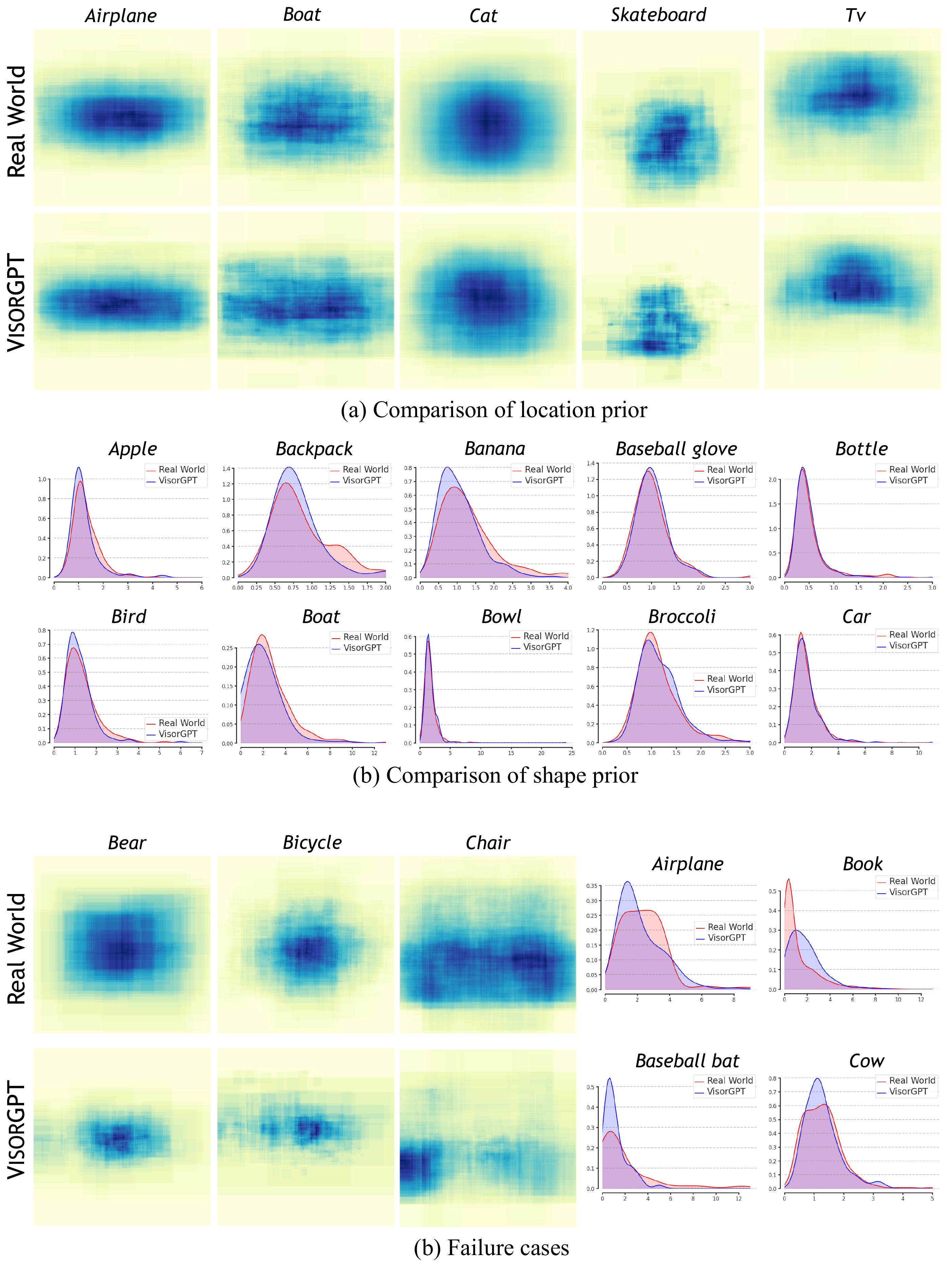}
		\caption{Comparison of visual prior between the real world and one learned by \our~on COCO dataset.} 
		\label{fig:prior_vis_appendix}
	\end{figure*}

    \textbf{More Visual Comparison}. We provide more comparison of visual prior between the real world and one learned by our \our~ and failure cases on COCO dataset in Fig.~\ref{fig:prior_vis_appendix}.

    \begin{figure*}[t]
		\centering
		\includegraphics[width=\linewidth]{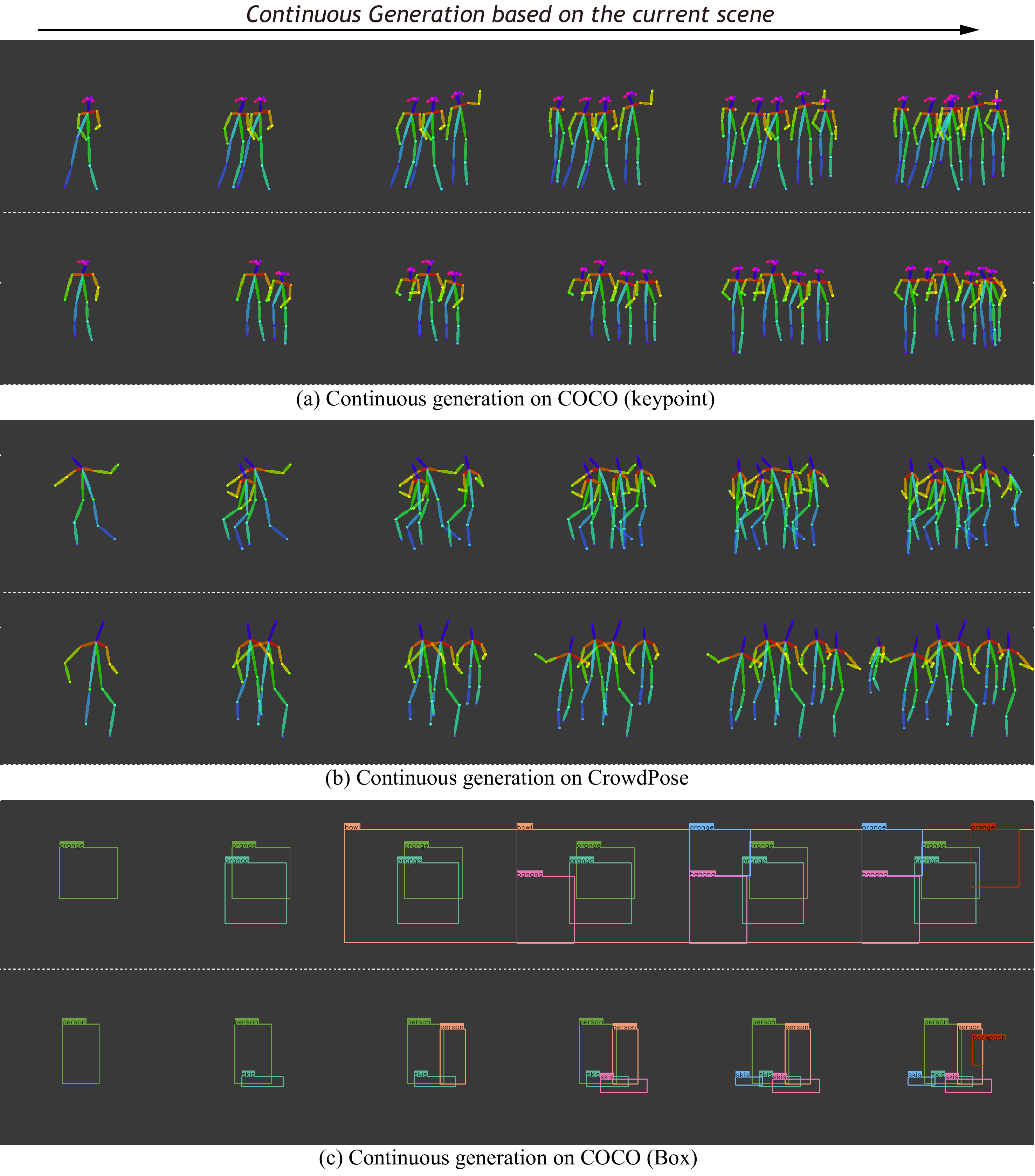}
            \vspace{-10pt}
		\caption{Examples of continual generation.} 
	    \vspace{-10pt}	
            \label{fig:cgen}
	\end{figure*}
    
    \textbf{Continuous Generation}. Fig.~~\ref{fig:cgen} presents a set of examples showcasing continuous generation based on the current scene. Notably, in each row, the proposed \our~is able to successfully complete a scene that involves many individuals annotated with 14/18 keypoints or objects with bounding boxes, based on the information provided in the corresponding scene depicted in the previous columns.

\begin{figure*}[t]
		\centering
		\includegraphics[width=\linewidth]{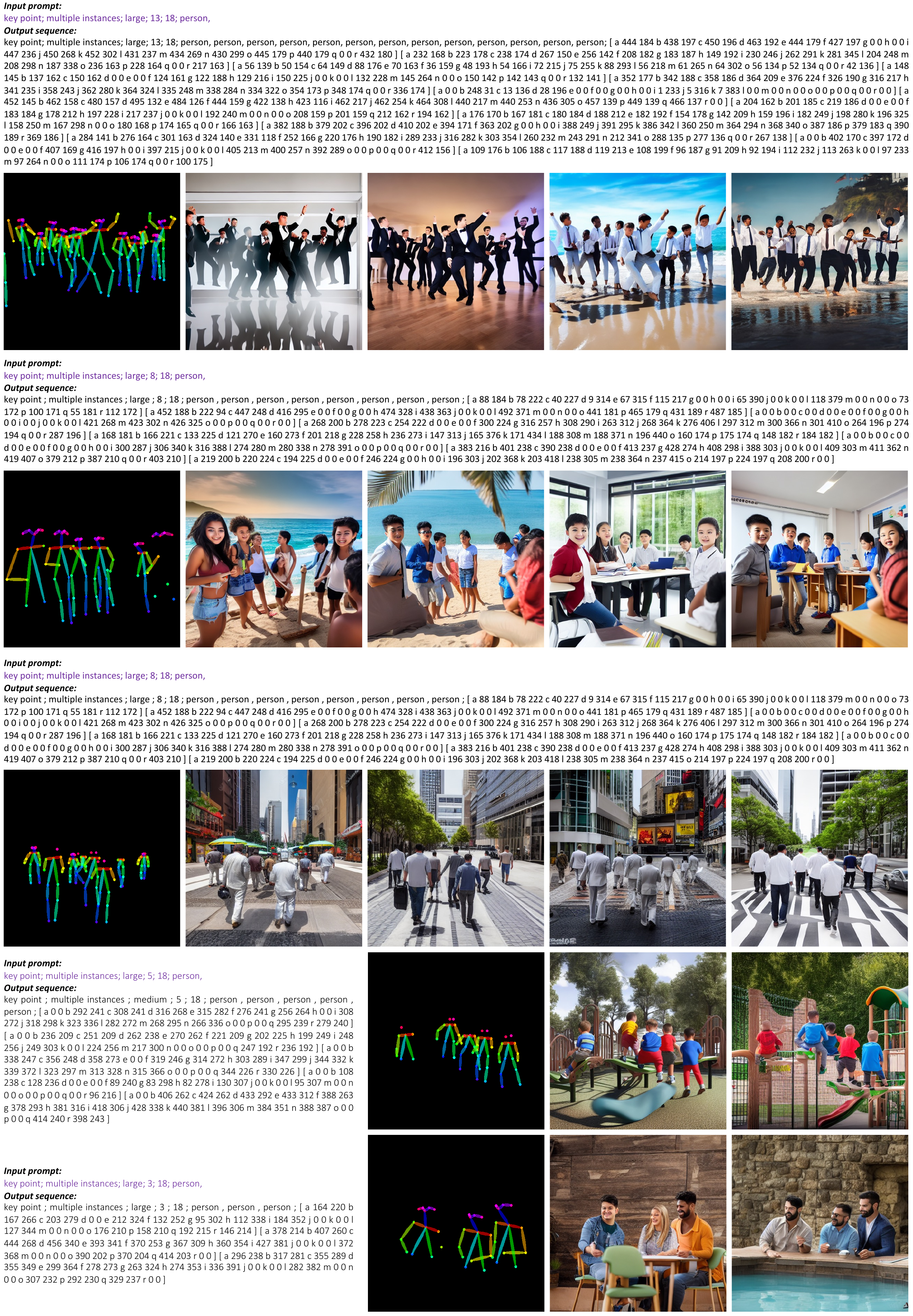}
            \vspace{-10pt}
		\caption{Examples of input prompts, output sequences, decoded results and synthetic images.} 
	    \vspace{-10pt}	
            \label{fig:seq2kpt2image_appendix}
	\end{figure*}

\begin{figure*}[t]
		\centering
		\includegraphics[width=\linewidth]{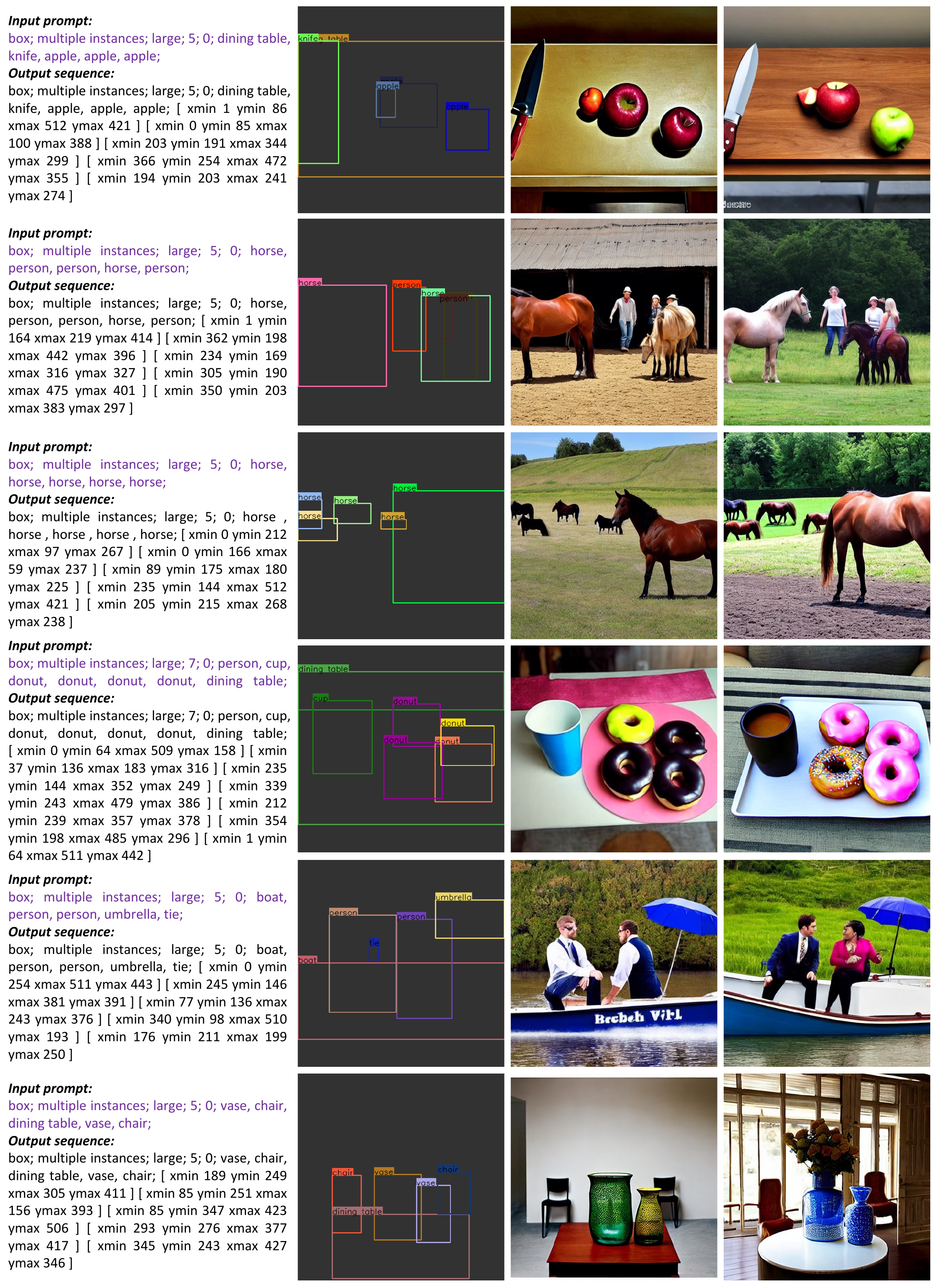}
            \vspace{-10pt}
		\caption{Examples of input prompts, output sequences, decoded results, and synthetic images.} 
	    \vspace{-10pt}	
            \label{fig:seq2box2image_appendix}
	\end{figure*}

    Figs.~\ref{fig:seq2kpt2image_appendix} and \ref{fig:seq2box2image_appendix} present more visualization results.


\end{document}